\documentclass{article}
\usepackage{spconf,amsmath,graphicx,hyperref,booktabs,array, threeparttable,makecell,amssymb,tabularx}
\usepackage{adjustbox}
\usepackage{arydshln}
\usepackage{array}
\usepackage[table]{xcolor} 
\usepackage{colortbl} 

\title{AFD-SLU: Adaptive Feature Distillation for Spoken Language Understanding}

\name{Yan Xie$^{1,2,3}$ \qquad Yibo Cui$^{2,3\ast}$ \qquad Liang Xie$^{2,3}$  \qquad Erwei Yin$^{2,3}$
\thanks{* Corresponding author.}
\thanks{This work was supported in part by the grants from the National Natural Science Foundation of China under Grant 62332019, the National Key Research and Development Program of China (2023YFF1203900, 2023YFF1203903), sponsored by Beijing Nova Program (20240484513).}
}
\address{$^{1}$ School of Advanced Manufacturing and Robotics, Peking University, Beijing, China \\
    $^{2}$ National Institute of Defense Technology Innovation, Academy of Military Sciences, Beijing, China  \\
    $^{3}$ Tianjin Artificial Intelligence Innovation Center (TAIIC), Tianjin, China}

\begin{document}
%
\maketitle
\begin{abstract}
Spoken Language Understanding (SLU) is a core component of conversational systems, enabling machines to interpret user utterances. Despite its importance, developing effective SLU systems remains challenging due to the scarcity of labeled training data and the computational burden of deploying Large Language Models (LLMs) in real-world applications. To further alleviate these issues, we propose an Adaptive Feature Distillation framework that transfers rich semantic representations from a General Text Embeddings (GTE)-based teacher model to a lightweight student model. Our method introduces a dynamic adapter equipped with a Residual Projection Neural Network (RPNN) to align heterogeneous feature spaces, and a Dynamic Distillation Coefficient (DDC) that adaptively modulates the distillation strength based on real-time feedback from intent and slot prediction performance. Experiments on the Chinese profile-based ProSLU benchmark demonstrate that AFD-SLU achieves state-of-the-art results, with 95.67\% intent accuracy, 92.02\% slot F1 score, and 85.50\% overall accuracy. 
\end{abstract}


\begin{figure*}[t] 
    \centering
    \includegraphics[trim=0.2cm 3cm 0.2cm 2cm, clip, width=0.9\linewidth]{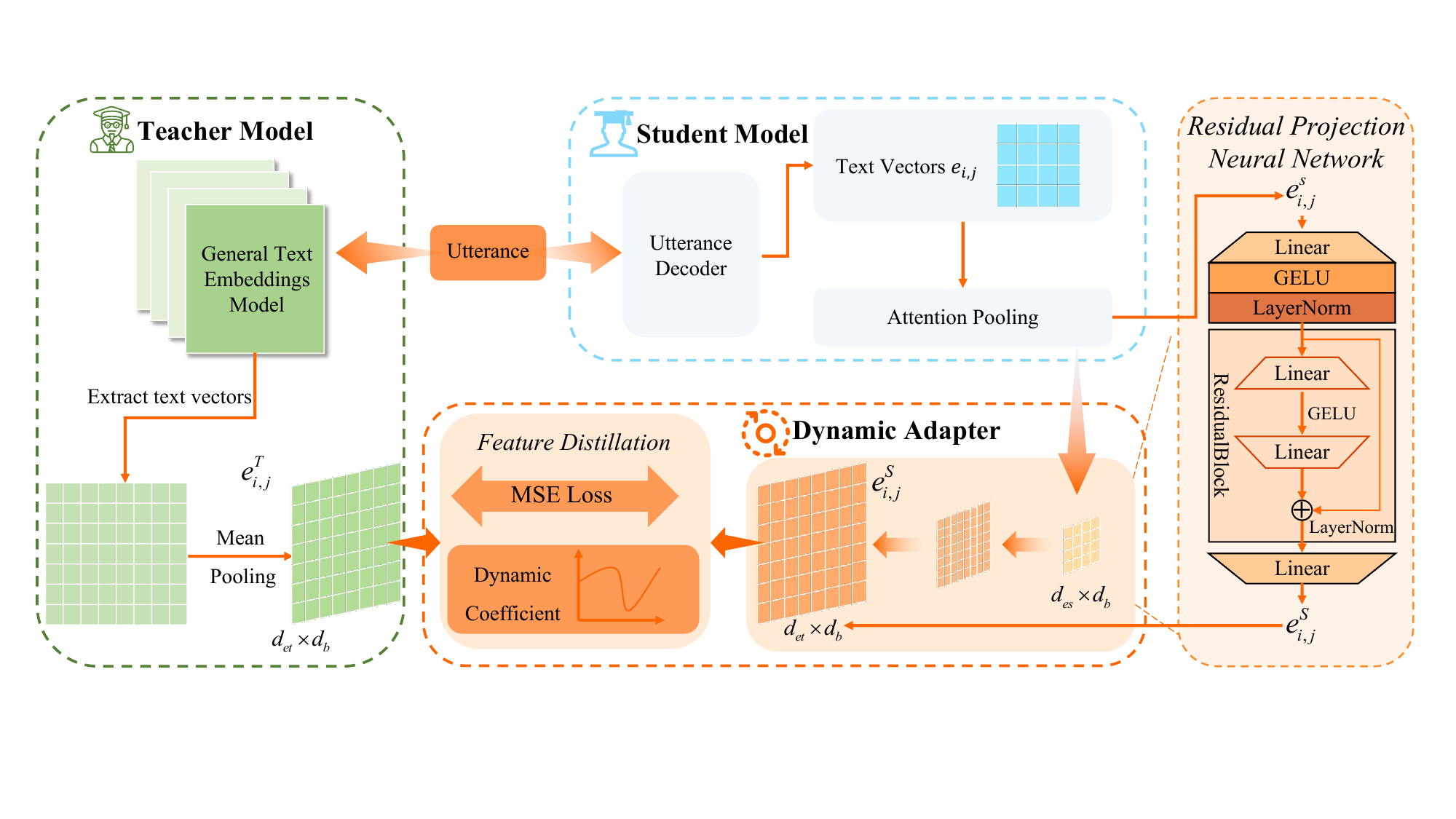} 
    \caption{\centering An overview of our model AFD-SLU.} 
    \label{fig:figure1} 
\end{figure*}

\section{Introduction}
\label{sec:intro}
Spoken Language Understanding (SLU) aims to transform human utterances into structured semantic representations and typically involves two core and closely related tasks: intent detection and slot filling \cite{NLUSurvey}. As a fundamental component of Natural Language Processing (NLP), SLU supports a broad range of applications, such as voice assistants, customer service systems, and smart devices. 

Early research in SLU primarily focused on capturing the inherent correlation between intent detection and slot filling \cite{SF-ID, Slot-Gated, Bi-model, AGIF, Stack-Propagation}. To further enhance model performance, subsequent work has explored knowledge distillation techniques leveraging pre-trained language models  \cite{NLUmethodssurvey}. Li et al. \cite{MIDAS} proposed a multi-teacher framework that transfers knowledge from multiple pre-trained models under limited supervision. 
However, task-level distillation are tailored to specific domains or languages, exhibiting limited cross-domain adaptability and generalization.
With the rapid advancement of Large Language Models (LLMs), recent studies have begun to harness their strong generalization and few-shot capabilities for SLU \cite{Zero-shotLLMSLU, zeroLLM, DC-Instruct}. Notably, Yin et al. \cite{ECLM} reformulated slot filling as an entity recognition task and proposes a "Chain of Intent" mechanism to support multi-intent understanding within an LLM-based framework. Despite their promising performance, LLMs entail significant memory and computational costs, which poses challenges for their deployment in real-world, resource-constrained SLU applications \cite{LLMEvaluation, LLM-Evaluating}.

These challenges are also prevalent in the Chinese. To address data scarcity, Xu et al. \cite{ProSLU} introduced ProSLU, a high-quality profile-based benchmark designed to address data scarcity. Building upon this resource, recent studies have explored the integration of user profile information to improve SLU performance. Thinh et al. \cite{JPIS} proposed a joint model with slot-to-intent attention, while Teng et al. \cite{PRO-HAN} developed a heterogeneous graph attention network that leverages profile features. Although these models perform well on ProSLU benchmark, they still encounter performance bottlenecks due to the dataset's limited scale.

To address these limitations, we propose \textbf{A}daptive \textbf{F}eature \textbf{D}istillation for Spoken Language Understanding (\textbf{AFD-SLU}), a lightweight framework for dynamic feature knowledge distillation. It comprises three components: a teacher model, a student model, and a dynamic adapter. The teacher model is a General Text Embeddings (GTE) model \cite{GTE} derived from a high-performance LLM, which provides generalizable semantic features while avoiding the heavy computational cost of direct LLM inference. To the best of our knowledge, this is the first work to employ GTE as a teacher model for SLU. The student model adopts a joint architecture for intent detection and slot filling, while the dynamic adapter integrates a Residual Projection Neural Network (RPNN) to align feature spaces and a Dynamic Distillation Coefficient (DDC) to adaptively adjust the distillation weight. 
Experiments on the ProSLU benchmark demonstrate that AFD-SLU achieves state-of-the-art performance.
Ablation studies verify the effectiveness of each module, and further experiments show the adaptability of our framework to different teacher models. All codes for this work are publicly available at https://github.com/XieyRock/AFD-SLU.

\section{APPROACH}
\label{sec:format}

This section presents the architecture of our \textbf{AFD-SLU}, as shown in Figure~\ref{fig:figure1}. The framework consists of three main components: a teacher model that provides token-level semantic embeddings (§2.1), a student model for joint intent detection and slot filling (§2.2), and a dynamic adapter for feature alignment and adaptive training (§2.3).

\subsection{Student Model}
\label{ssec:Student Model}
 The student model adopts a joint SLU architecture designed for Chinese linguistic characteristics and optimized for efficient inference. In this work, we adopt a Bidirectional Long Short-Term Memory (BiLSTM)-based joint SLU model \cite{JPIS, PRO-HAN} as the student model, which is currently one of the most commonly used architectures for joint models. Given an input sequence $X=\{{{x}_{1}},{{x}_{2}},\cdots,{{x}_{n}}\}$  with n tokens, a BiLSTM followed by a self-attention layer is applied to capture sequential and contextual information. An attention pooling layer is then applied to aggregate these features, yielding a representation $e_{i,j}^{s} \in \mathbb{R}^{d_{b} \times d_{es}}$, where $d_{b}$ is the batch size and $d_{es}$ denotes the dimensionality of the student model’s text embeddings. Here, $i$ refers to the $i$-th sample in the batch and $j$ denotes the $j$-th token in the sequence.

\subsection{Teacher Model}
\label{ssec:Teacher Model}

We adopt the GTE model \cite{GTE} as the teacher model. During the entire training process, the teacher model is solely responsible for generating sentence embeddings and providing supervision signals to the student model, with its parameters kept frozen. The generation of sentence embeddings involves the following three steps:

\textbf{Utterance Encoding}: Given a user utterance $X = \{x_1, x_2, \cdots, x_n\}$, identical to the one used in the student model, it is first tokenized into a sequence $X \in \mathbb{Z}^L$ using a pre-trained tokenizer. To conform to the input constraints of the student model, the token sequence is padded to a fixed length $L$, resulting in a standardized input tensor. Meanwhile, an attention mask $m \in \{0,1\}^L$ is constructed using a standard masking algorithm to indicate valid token positions.

\textbf{Feature Extraction}: We extract the outputs from the last four hidden layers of the teacher model, denoted as $H = [h_1, h_2, h_3, h_4] \in \mathbb{R}^{4 \times L \times d_{et}}$, where $d_{et}$ is the dimensionality of the teacher model’s text embeddings. Each $h_k$ corresponds to the hidden representation from the $k$-th last layer.

\textbf{Embedding Generation}: The four hidden states are averaged to obtain a unified feature representation. This representation is then passed through mean pooling over the valid tokens (as indicated by the attention mask) to derive the final sentence embeddings. The computation is defined as:

\begin{equation} \label{eq1}
e_{i,j}^{T} = \frac{\sum\limits_{l=1}^{L} \bar{h}_{l} \cdot m_l}{\sum\limits_{l=1}^{L} m_l}
\end{equation}
Here, $\bar{h}_l$ represents the averaged hidden state at position $l$ across the last four layers of the teacher model. The representation $e_{i,j}^T \in \mathbb{R}^{d_b \times d_{et}}$ denotes the teacher model's embeddings of the $j$-th token in the $i$-th sample.

\subsection{Dynamic Adapter}
\label{ssec:Dynamic Adapter}

\subsubsection{Residual Projection Neural Network}
\label{sssec:Residual Projection Neural Network}

In our model, the Residual Projection Neural Network (RPNN) plays a critical role in aligning feature representations between the teacher and student model. Its primary goal is to project the lower-dimensional embedding space of the student model into the higher-dimensional space of the teacher model through a nonlinear transformation, while preserving essential information from the original features.

Let the text embeddings output from the student model be denoted as $e_{i,j}^{s} \in \mathbb{R}^{d_b \times d_{es}}$, and the corresponding embeddings from the teacher model as $e_{i,j}^{T} \in \mathbb{R}^{d_b \times d_{et}}$, where $d_{es} \ne d_{et}$. The RPNN first projects $e_{i,j}^s$ into a higher-dimensional hidden space using a linear layer, followed by a GELU activation and Layer Normalization (LN):
\begin{equation}\label{eq2}
h_1 = \text{LN}_1\left( \text{GELU}(W_1 e_{i,j}^s + b_1) \right)
\end{equation}
where $W_1 \in \mathbb{R}^{d_{es} \times d_h}$, $b_1 \in \mathbb{R}^{d_h}$, and $d_h = 4d_{es}$.

To further enhance representational capacity, we introduce a residual transformation module, denoted as \textit{ResidualBlock}, which is shown on the right of Figure~\ref{fig:figure1}. This module refines $h_1$ through a two-layer feed-forward network with a skip connection, followed by another layer normalization:
\begin{equation}\label{eq3}
h_2 = \text{LN}_2\left( h_1 + W_3 \cdot \text{GELU}(W_2 h_1 + b_2) + b_3 \right)
\end{equation}
where $W_2 \in \mathbb{R}^{d_h \times d_h}$, $W_3 \in \mathbb{R}^{d_h \times d_{es}}$, $b_2 \in \mathbb{R}^{d_h}$, and $b_3 \in \mathbb{R}^{d_{es}}$. The residual connection facilitates gradient flow and preserves information from earlier layers.

Finally, the refined features are projected into the teacher model’s embeddings space:
\begin{equation}\label{eq4}
e_{i,j}^{S} = W_4 h_2 + b_4
\end{equation}
where $W_4 \in \mathbb{R}^{d_{es} \times d_{et}}$ and $b_4 \in \mathbb{R}^{d_{et}}$.

\subsubsection{Dynamic Distillation Coefficient}
\label{sssec:Dynamic Distillation Coefficient}

Fixed-weight strategies often fail to balance the distillation loss and the task loss effectively. They cannot reflect the changing learning capacity of the student model during training. This limits the efficiency of knowledge transfer.

We introduce a Dynamic Distillation Coefficient (DDC) that adaptively balances distillation and task losses during training. The distillation loss is computed as the Mean Squared Error (MSE) between the token-level embeddings of the teacher and student models:
\begin{equation} \label{eq5}
\mathcal{L}_{\text{distill}} = \frac{1}{d_b} \sum_{i=1}^{d_b} \sum_{j=1}^{d_{et}} \left( e_{i,j}^{T} - e_{i,j}^{S} \right)^2
\end{equation} 

The dynamic coefficient $\lambda$ modulates the relative contribution of the distillation and task losses throughout training:
\begin{equation} \label{eq6}
\lambda = \lambda_{\text{final}} + (\lambda_{\text{initial}} - \lambda_{\text{final}})\left(1 + \cos\left(\frac{e}{E} \pi\right)\right)
\end{equation}
where $\lambda_{\text{initial}}$ and $\lambda_{\text{final}}$ denote the initial and final weights, $e$ is the current epoch, and $E$ is the total number of epochs.
This cosine annealing schedule gradually reduces the influence of the teacher model, enabling the student model to increasingly focus on task-specific learning in the later stages of training.

\begin{equation} \label{eq7}
\mathcal{L}_{\text{task}} = \alpha \mathcal{L}_{\text{ID}} +(1-\alpha) \mathcal{L}_{\text{SF}}
\end{equation}
where $\mathcal{L}_{\text{ID}}$ and $\mathcal{L}_{\text{SF}}$ denote the intent loss and the slot loss.

The overall training objective is formulated as:
\begin{equation} \label{eq8}
\mathcal{L}_{\text{total}} = \mathcal{L}_{\text{task}} + \lambda \mathcal{L}_{\text{distill}}
\end{equation}

\section{EXPERIMENTS}
\label{sec:pagestyle}

\subsection{Experimental Setup}
\label{ssec:Dataset and Evaluation metrics}
\textbf{Dataset and Evaluation metrics.} 
To evaluate the effectiveness of the proposed framework, we conduct experiments on the ProSLU dataset, a Chinese spoken language understanding benchmark that incorporates user profile information. The dataset contains 4,196 training utterances, 522 validation utterances, and 531 test utterances. Each utterance is accompanied by four user profile vectors and four context-aware vectors, and the annotations cover 14 intents and 99 slots. Model performance is assessed using three standard metrics: slot F1 score, intent accuracy, and overall accuracy.

\noindent\textbf{Implementation Details.} 
We train our model on a NVIDIA RTX 4090D GPU equipped with 48 GB of memory. We optimize the model using the Adam and train it for 50 epochs, applying a dropout rate of 0.4. 
The dynamic parameter $\lambda$ was selected based on comparative validation. Its initial and final values were set to $\lambda_{\text{initial}} = 0.1$ and $\lambda_{\text{final}} = 0.7$, respectively.
We evaluate our method with both JPIS \cite{JPIS} and PRO-HAN \cite{PRO-HAN} serving individually as student models, and employ the gte\_Qwen2-1.5B-instruct \cite{GTE} as teacher model. For JPIS, we use a batch size of 1 and a learning rate of \(2 \times 10^{-4}\); for PRO-HAN, the batch size is 64 and the learning rate is \(1 \times 10^{-3}\).

\subsection{Main Results}
\label{ssec:Main results}

We apply our AFD-SLU framework to the above student–teacher settings. As shown in Table~\ref{tab:main-results}, integrating AFD improves performance across all evaluation metrics. For JPIS, applying AFD yields absolute gains of 2.25\% in intent accuracy, 2.75\% in slot F1 score, and 3.20\% in overall accuracy. Similarly, PRO-HAN-AFD achieves 95.67\% intent accuracy and 92.02\% slot F1 score, outperforming the original model by 3.58\% and 2.01\%, respectively, while maintaining comparable overall accuracy. These results demonstrate that AFD-SLU significantly enhances model performance under low-resource conditions, while keeping the model parameters and the training data scale nearly unchanged. The consistent gains confirm that our method effectively transfers semantic knowledge from GTE models to student models, improving both intent detection and slot filling without increasing inference overhead.

\begin{table}[htbp]
    \centering
    \small
    \setlength{\tabcolsep}{4pt} 
    \renewcommand{\arraystretch}{1.1}
    \caption{Main results on the ProSLU dataset. Models with ``+AFD'' denote variants enhanced with adaptive feature distillation. 
    $^{\dagger}$ indicates our reproduced results.
    }
    \label{tab:main-results}
    \begin{adjustbox}{max width=\columnwidth}
    \begin{tabular}{lcccc}
        \toprule
        \textbf{Model} & \textbf{Intent (Acc)} & \textbf{Slot (F1)} & \textbf{Overall (Acc)} & \textbf{Params} \\
        \midrule
        SF-ID \cite{SF-ID} & 83.24 & 73.70 & 68.36 & $<$1M \\
        Slot-Gated \cite{Slot-Gated} & 83.24 & 74.18 & 69.11 & $<$1M \\
        Bi-Model \cite{Bi-model} & 81.92 & 77.76 & 73.45 & $\sim$10M \\
        AGIF \cite{AGIF} & 81.54 & 80.57 & 74.95 & $\sim$1M \\
        Stack-Propagation \cite{Stack-Propagation} & 83.99 & 81.08 & 78.91 & $\sim$10M \\
        General-SLU \cite{ProSLU} & 85.31 & 83.27 & 79.10 & $\sim$1M \\
        GL-GIN \cite{GL-GIN} & 85.69 & 82.70 & 79.28 & $\sim$1M \\
        \midrule
        JPIS \cite{JPIS} & 87.95 & 85.76 & 82.30 & $\sim$2M \\
        \textbf{JPIS-AFD (Ours)} & \textbf{90.20} & \textbf{88.51} & \textbf{85.50} & $\sim$2M \\
        \midrule
        PRO-HAN$^{\dagger}$ \cite{PRO-HAN} & 92.09 & 90.01 & 85.31 & $\sim$40M \\
        \textbf{PRO-HAN-AFD (Ours)} & \textbf{95.67} & \textbf{92.02} & \textbf{85.50} & $\sim$40M \\
        \bottomrule
    \end{tabular}
    \end{adjustbox}
\end{table}

\begin{table}[htbp]
    \centering
    \caption{Ablation study results using JPIS as the student model and gte\_Qwen2-1.5B-instruct as the teacher model. }
    \label{tab:ablation_results}
    \begin{adjustbox}{width=1\columnwidth}
    \setlength{\tabcolsep}{6pt}
    \begin{tabular}{l:ccc}
        \toprule
        \textbf{Model} & \textbf{Intent (Acc)} & \textbf{Slot (F1)} & \textbf{Overall (Acc)} \\
        \midrule
         \textbf{JPIS-AFD}& \textbf{90.20} & \textbf{88.51} & \textbf{85.50} \\
        \textit{RPNN(Linear)} & 87.95 & 86.13 & 83.43 \\
        \textit{RPNN(Deep)} & 88.70 & 87.13 & 83.99 \\
        \textit{w/o DDC}  & 87.59 & 85.96 & 83.24 \\
        \bottomrule
    \end{tabular}
    \end{adjustbox}
\end{table}

\subsection{Ablation study}
\label{ssec:Ablation study}
We perform an ablation study to assess the contributions of key components in our framework. 
As shown in Table~\ref{tab:ablation_results}, \textit{RPNN (Linear)} applies a single linear projection and results in performance drops of 2.01\% in intent accuracy, 2.38\% in slot F1 score, and 2.07\% in overall accuracy compared to the JPIS-AFD model. Meanwhile, \textit{RPNN (Deep)} introduces a deeper nonlinear projection network, also leads to reduced performance.This observation suggests that effective feature alignment benefits from a carefully balanced RPNN design, whereas both overly simple linear transformations and excessively complex projection networks are suboptimal.
Similarly, excluding the DDC leads to reductions of 2.61\%, 2.55\%, and 2.26\% in the same metrics. This highlights the importance of adaptively adjusting the distillation weight to balance the student’s intrinsic learning and the knowledge transferred from the teacher during training. 
These results confirm that both RPNN and DDC are essential to the effectiveness of our method, working together to significantly enhance performance on joint SLU tasks.

\subsection{Analysis}
\label{ssec:Analysis}
We evaluated the impact of different GTE models used as teacher models, with results shown in Table \ref{tab:model_influence}. We can conclude that: Smaller or moderately sized teacher models that are well-aligned with the task, such as GTE-Qwen2-1.5B-instruct and Qwen3-Embedding-0.6B, are more effective in guiding the student model through knowledge distillation. These models achieve better performance across intent recognition, slot filling, and overall accuracy. In contrast, overly large models tend to overfit when trained on small datasets such as ProSLU, exhibiting high initial accuracy but subsequently suffering from overfitting, which leads to suboptimal generalization and degraded performance. Additionally, pre-trained models such as GTE-large-zh may lack sufficient task-specific understanding, which limits their effectiveness as teacher models. These findings emphasize the importance of selecting a teacher model that is not only powerful but also well-suited to the task and dataset scale.


\begin{table}[htbp]
    \centering
    \small
    \setlength{\tabcolsep}{6pt}
    \renewcommand{\arraystretch}{1.1}
    \caption{Impact of different general text embedding models used as teacher models on knowledge distillation performance, with JPIS as the student model.}
    \label{tab:model_influence}
    \begin{adjustbox}{max width=\columnwidth}
    \begin{tabular}{lccc}
        \toprule
        \textbf{Teacher Model} & \textbf{Intent (Acc)} & \textbf{Slot (F1)} & \textbf{Overall (Acc)} \\
        \midrule
        gte\_large-zh & 88.14 & 86.11 & 83.24 \\
        gte\_Qwen2-1.5B-instruct & 90.20 & \textbf{88.51} & \textbf{85.50} \\
        gte\_Qwen2-7B-instruct & 89.64 & 88.00 & 84.75 \\
        Qwen3-Embedding-0.6B & \textbf{90.21} & 88.47 & 85.12 \\
        Qwen3-Embedding-4B & 88.14 & 86.80 & 84.18 \\
        Qwen3-Embedding-8B & 87.01 & 85.27 & 82.30 \\
        \bottomrule
    \end{tabular}
    \end{adjustbox}
\end{table}

\section{CONCLUSION}
\label{sec:CONCLUSION}
In this paper, we propose AFD-SLU, an Adaptive Feature Distillation framework aimed at enhancing Chinese SLU in low-resource scenarios. AFD-SLU leverages a powerful GTE model as a teacher to transfer extensive semantic knowledge to a lightweight student model. Extensive experiments on the ProSLU benchmark demonstrate that AFD-SLU achieves state-of-the-art performance, significantly outperforming baselines in intent detection, slot filling, and overall accuracy. Ablation studies validate the contribution of each component, and comparisons across teacher models reveal that models with larger parameter sizes do not necessarily lead to better student performance; selecting a teacher suited to the task is crucial. 
Future work may explore more effective data augmentation and model compression to further enhance practicality in real-world low-resource scenarios.



\vfill\pagebreak


\bibliographystyle{IEEEbib}
\bibliography{strings,refs}

@inproceedings{zeroLLM,
    title = "Zero-shot Slot Filling in the Age of {LLM}s for Dialogue Systems",
    author = "Rana, Mansi  and
      Hacioglu, Kadri  and
      Gopalan, Sindhuja  and
      Boothalingam, Maragathamani",
    booktitle = "Proc. of COLING",
    year = "2025",
    pages = "697--706",
}

@INPROCEEDINGS{ProSLU,
  title={Text is no more enough! a benchmark for profile-based spoken language understanding},
  author={Xu, Xiao and Qin, Libo and Chen, Kaiji and Wu, Guoxing and Li, Linlin and Che, Wanxiang},
  booktitle={Proc. of AAAI},
  volume={36},
  number={10},
  pages={11575--11585},
  year={2022}
}

@INPROCEEDINGS{JPIS,
  author={Pham, Thinh and Nguyen, Dat Quoc},
  booktitle={Proc. of ICASSP}, 
  organization = "IEEE",
  title={{JPIS}: A Joint Model for Profile-Based Intent Detection and Slot Filling with Slot-to-Intent Attention}, 
  year={2024},
  pages={10446-10450},
}

@INPROCEEDINGS{PRO-HAN,
  author={Teng, Dechuan and Lu, Chunlin and Xu, Xiao and Che, Wanxiang and Qin, Libo},
  booktitle={Proc. of ICASSP}, 
  organization = "IEEE",
  title={{PRO-HAN}: A Heterogeneous Graph Attention Network for Profile-based Spoken Language Understanding}, 
  year={2024},
  pages={10726-10730},
}

@inproceedings{SF-ID,
    title = "A Novel Bi-directional Interrelated Model for Joint Intent Detection and Slot Filling",
    author = "E, Haihong  and
      Niu, Peiqing  and
      Chen, Zhongfu  and
      Song, Meina",
    booktitle = "Proc. of ACL",
    year = "2019",
    pages = "5467--5471"
}

@inproceedings{Slot-Gated,
    title = "Slot-Gated Modeling for Joint Slot Filling and Intent Prediction",
    author = "Goo, Chih-Wen  and
      Gao, Guang  and
      Hsu, Yun-Kai  and
      Huo, Chih-Li  and
      Chen, Tsung-Chieh  and
      Hsu, Keng-Wei  and
      Chen, Yun-Nung",
    booktitle = "Proc. of NAACL",
    volume={2},
    year = "2018",
    pages = "753--757"
}

@inproceedings{Bi-model,
    title = "A Bi-Model Based {RNN} Semantic Frame Parsing Model for Intent Detection and Slot Filling",
    author = "Wang, Yu  and
      Shen, Yilin  and
      Jin, Hongxia",
    booktitle = "Proc. of NAACL",
    volume={2},
    year = "2018",
    pages = "309--314"
}

@inproceedings{AGIF,
    title = "{AGIF}: An Adaptive Graph-Interactive Framework for Joint Multiple Intent Detection and Slot Filling",
    author = "Qin, Libo  and
      Xu, Xiao  and
      Che, Wanxiang  and
      Liu, Ting",
    booktitle = "Proc. of EMNLP",
    year = "2020",
    pages = "1807--1816"
}

@inproceedings{Stack-Propagation,
    title = "A Stack-Propagation Framework with Token-Level Intent Detection for Spoken Language Understanding",
    author = "Qin, Libo  and
      Che, Wanxiang  and
      Li, Yangming  and
      Wen, Haoyang  and
      Liu, Ting",
    booktitle = "Proc. of EMNLP",
    year = "2019",
    pages = "2078--2087"
}

@inproceedings{GL-GIN,
    title = "{GL}-{GIN}: Fast and Accurate Non-Autoregressive Model for Joint Multiple Intent Detection and Slot Filling",
    author = "Qin, Libo  and
      Wei, Fuxuan  and
      Xie, Tianbao  and
      Xu, Xiao  and
      Che, Wanxiang  and
      Liu, Ting",
    booktitle = "Proc. of ACL",
    volume={1},
    year = "2021",
    pages = "178--188"
}

@inproceedings{MIDAS,
    title = "{MIDAS}: Multi-level Intent, Domain, And Slot Knowledge Distillation for Multi-turn {NLU}",
    author = "Li, Yan  and
      Kim, So-Eon  and
      Park, Seong-Bae  and
      Han, Caren",
    booktitle = "Proc. of NAACL",
    year = "2025",
    pages = "7989--8012",
}

@inproceedings{LLM-Evaluating,
    title = "A Systematic Survey and Critical Review on Evaluating Large Language Models: Challenges, Limitations, and Recommendations",
    author = "Laskar, Md Tahmid Rahman  and
      Alqahtani, Sawsan  and
      Bari, M Saiful  and
      Rahman, Mizanur  and
      Khan, Mohammad Abdullah Matin  and
      Khan, Haidar  and
      Jahan, Israt  and
      Bhuiyan, Amran  and
      Tan, Chee Wei  and
      Parvez, Md Rizwan  and
      Hoque, Enamul  and
      Joty, Shafiq  and
      Huang, Jimmy",
    booktitle = "Proc. of EMNLP",
    year = "2024",
    pages = "13785--13816",
}

@inproceedings{ECLM,
  author       = {Shangjian Yin and
                  Peijie Huang and
                  Jiatian Chen and
                  Haojing Huang and
                  Yuhong Xu},
  title        = {{ECLM:} Entity Level Language Model for Spoken Language Understanding
                  with Chain of Intent},
  booktitle    = {Proc. of ACL},
  pages        = {21851--21862},
  year         = {2025},
}

@inproceedings{Zero-shotLLMSLU,
  author       = {Zhihong Zhu and
                  Xuxin Cheng and
                  Hao An and
                  Zhichang Wang and
                  Dongsheng Chen and
                  Zhiqi Huang},
  title        = {Zero-Shot Spoken Language Understanding via Large Language Models:
                  {A} Preliminary Study},
  booktitle    = {Proc. of LREC/COLING},
  pages        = {17877--17883},
  year         = {2024},
}

@inproceedings{DC-Instruct,
    title = "{DC}-Instruct: An Effective Framework for Generative Multi-intent Spoken Language Understanding",
    author = "Xing, Bowen  and
      Liao, Lizi  and
      Huang, Minlie  and
      Tsang, Ivor",
    booktitle = "Proc. of EMNLP",
    year = "2024",
    pages = "14520--14534"
}

@article{NLUSurvey,
author = {Weld, Henry and Huang, Xiaoqi and Long, Siqu and Poon, Josiah and Han, Soyeon Caren},
title = {A Survey of Joint Intent Detection and Slot Filling Models in Natural Language Understanding},
year = {2022},
volume = {55},
number = {8},
journal = {ACM Comput. Surv.},
articleno = {156},
numpages = {38},
Pages= "1 -- 38"
}

@article{LLMEvaluation,
author = {Chang, Yupeng and Wang, Xu and Wang, Jindong and Wu, Yuan and Yang, Linyi and Zhu, Kaijie and Chen, Hao and Yi, Xiaoyuan and Wang, Cunxiang and Wang, Yidong and Ye, Wei and Zhang, Yue and Chang, Yi and Yu, Philip S. and Yang, Qiang and Xie, Xing},
title = {A Survey on Evaluation of Large Language Models},
year = {2024},
issue_date = {June 2024},
publisher = {Association for Computing Machinery},
address = {New York, NY, USA},
volume = {15},
number = {3},
issn = {2157-6904},
url = {https://doi.org/10.1145/3641289},
doi = {10.1145/3641289},
journal = {ACM Trans. Intell. Syst. Technol.},
articleno = {39},
numpages = {45},
pages = "1 -- 45"
}

@article{NLUmethodssurvey,
    title = "Efficient Methods for Natural Language Processing: A Survey",
    author = "Treviso, Marcos  and
      Lee, Ji-Ung  and
      Ji, Tianchu  and
      van Aken, Betty  and
      Cao, Qingqing  and
      Ciosici, Manuel R.  and
      Hassid, Michael  and
      Heafield, Kenneth  and
      Hooker, Sara  and
      Raffel, Colin  and
      Martins, Pedro H.  and
      Martins, Andr{\'e} F. T.  and
      Forde, Jessica Zosa  and
      Milder, Peter  and
      Simpson, Edwin  and
      Slonim, Noam  and
      Dodge, Jesse  and
      Strubell, Emma  and
      Balasubramanian, Niranjan  and
      Derczynski, Leon  and
      Gurevych, Iryna  and
      Schwartz, Roy",
    journal = "Transactions of the Association for Computational Linguistics",
    volume = "11",
    year = "2023",
    address = "Cambridge, MA",
    publisher = "MIT Press",
    url = "https://aclanthology.org/2023.tacl-1.48/",
    doi = "10.1162/tacl_a_00577",
    pages = "826--860",
}

@article{GTE,
    author = {Li, Zehan and Zhang, Xin and Zhang, Yanzhao and Long, Dingkun and Xie, Pengjun and Zhang, Meishan},
    title = {Towards General Text Embeddings with Multi-stage Contrastive Learning},
    journal = {arXiv preprint arXiv:2308.03281},
    year = {2023},
    primaryclass = {cs.CL},
    url = {https://arxiv.org/abs/2308.03281}
}

\end{document}